\documentclass[10pt,twocolumn,letterpaper]{article}

\usepackage{pifont}
\usepackage{multirow}
\usepackage{algorithm}
\usepackage{algpseudocode}
\usepackage[accsupp]{axessibility}  
\algrenewcommand\algorithmicrequire{\textbf{Require:}}
\algrenewcommand\algorithmicensure{\textbf{Ensure:}}
\newcommand{\GPO}{\operatorname{GPO}}

\usepackage[pagenumbers]{wacv} 

\definecolor{wacvblue}{rgb}{0.21,0.49,0.74}
\usepackage[pagebackref,breaklinks,colorlinks,allcolors=wacvblue]{hyperref}

\def\wacvPaperID{3393} 
\def\confName{WACV}
\def\confYear{2026}

\title{ITSELF: Attention Guided Fine-Grained Alignment \\ for Vision–Language Retrieval}

\author{
Tien-Huy Nguyen$^{1,2,*}$ \quad Huu-Loc Tran$^{1,2}$\thanks{Authors contributed equally to this paper.\\} \quad Thanh Duc Ngo$^{1,2}$ \\
$^{1}$ University of Information Technology, Ho Chi Minh City, VIETNAM \\
$^{2}$ Vietnam National University, Ho Chi Minh City, VIETNAM \\}

\begin{document}
\maketitle
\begin{abstract}

Vision Language Models (VLMs) have rapidly advanced and show strong promise for text-based person search (TBPS), a task that requires capturing fine-grained relationships between images and text to distinguish individuals. Previous methods address these challenges through local alignment, yet they are often prone to shortcut learning and spurious correlations, yielding misalignment. Moreover, injecting prior knowledge can distort intra-modality structure. Motivated by our finding that encoder attention surfaces spatially precise evidence from the earliest training epochs, \emph{and} to alleviate these issues, we introduce \textbf{ITSELF}, an attention-guided framework for \emph{implicit local alignment}. At its core, Guided Representation with Attentive Bank (GRAB) converts the model’s own attention into an Attentive Bank of high-saliency tokens and applies local objectives on this bank, learning fine-grained correspondences without extra supervision. To make the selection reliable and non-redundant, we introduce Multi-Layer Attention for Robust Selection (MARS), which aggregates attention across layers and performs diversity-aware top-k selection; and Adaptive Token Scheduler (ATS), which schedules the retention budget from coarse to fine over training, preserving context early while progressively focusing on discriminative details. Extensive experiments on three widely used TBPS benchmarks show \textbf{state-of-the-art} performance and strong cross-dataset generalization, confirming the effectiveness and robustness of our approach without additional prior supervision. Our project is publicly available at \textcolor{blue}{https://trhuuloc.github.io/itself}

\end{abstract}

\section{Introduction}
\label{sec:intro}


\begin{figure}[t]
    \centering
    \includegraphics[scale=0.3]{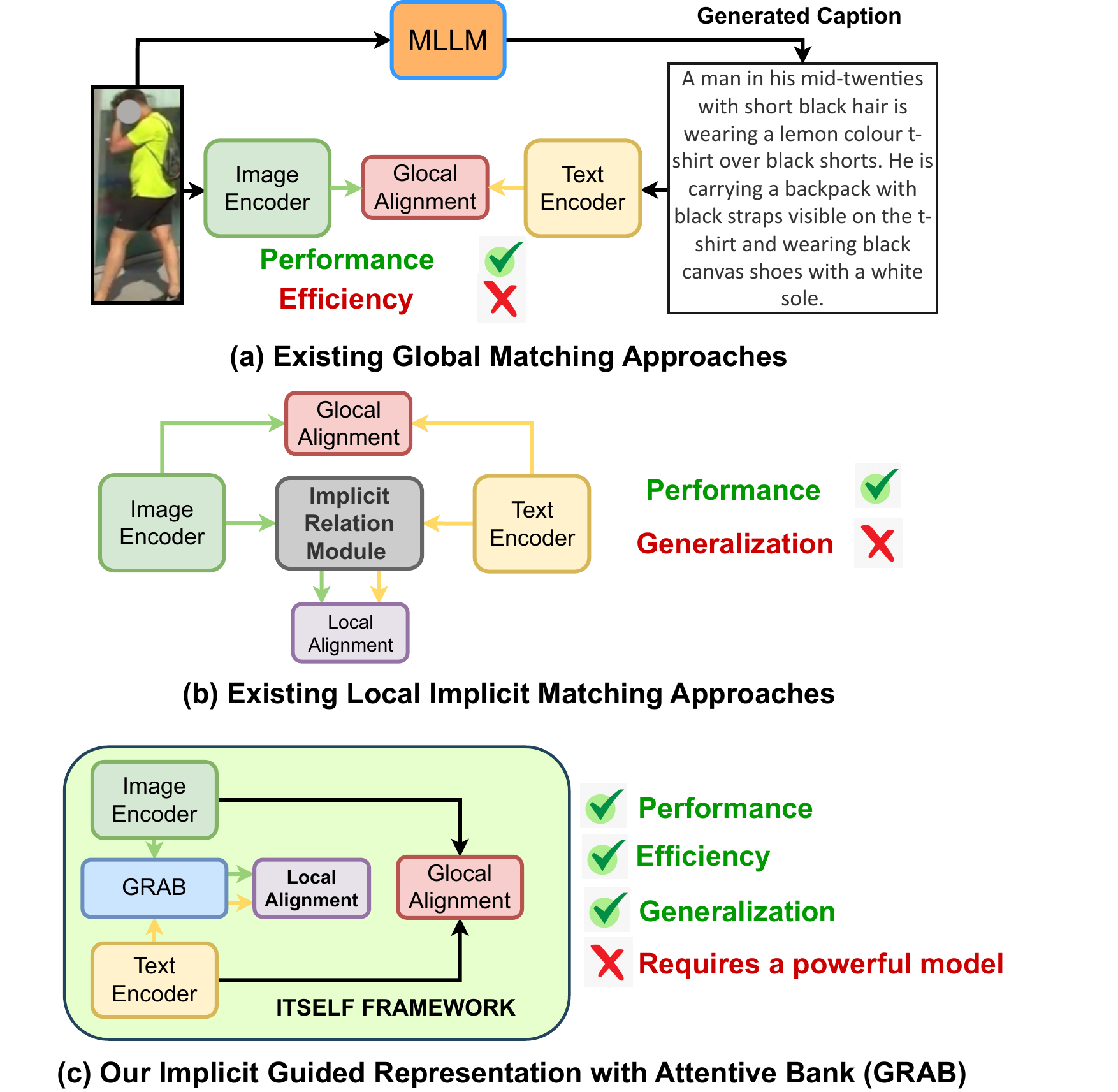}
    \caption{Evolution of text-based person search paradigms. (a) Global matching method uses powerful MLLM to synthesize extra datasets (b) Recent local implicit matching method implicitly reasons over relations among all local tokens. (c) Ours - ITSELF with GRAB: an attention-guided local branch to learn implicitly fine-grained, discriminative features to achieve better alignment.}
    \label{fig1:motivation}
\end{figure}

Text-based person search (TBPS) aims to identify, from a large image gallery, the person best matching a textual query \cite{cuhk}. Solving it requires extracting identity-discriminative cues from both image and text to distinguish individuals with subtle differences. Recent advances in vision language models (VLMs) \cite{li2025surveystateartlarge, zhang2024visionlanguagemodelsvisiontasks,nguyennhu2025stervlmspatiotemporalenhancedreference,tran2025efficientrobustmomentretrieval,nguyennhu2025lightweightmomentretrievalglobal}, notably CLIP \cite{clip}, have shown strong potential for tackling these fine-grained challenges. Building on this foundation, TBPS-CLIP \cite{tbpsclip} was the pioneer to apply CLIP to TBPS, followed by extensions \cite{irra, plot, cfam, dmadapter,yan2022clipdrivenfinegrainedtextimageperson,nguyen2025hybridunifiediterativenovel} that further narrow the text–image gap. However, many recent methods \cite{plip, Song_Hu_Zhao_2024, 10.1145/3664647.3681553,cfam,aptm,nguyen2024improvinggeneralizationvisualreasoning}  rely on costly external resources. For example, using MLLMs to synthesize auxiliary data (\textbf{\textcolor{red}{\cref{fig1:motivation}(a)}}), while effective, increase compute and annotation costs, and hinder scalability and robustness.

To motivate the limitation, we pose a simple question: \textbf{How can a TBPS model capture fine-grained, discriminative details on its own without costly external supervision?} To minimize this, recent work pursues \emph{implicit local alignment} (\textbf{\textcolor{red}{\cref{fig1:motivation}(b)}}) to explore discriminative cues without burdensome external supervision. Embedding-space correspondence methods \cite{axm, 10.1007/978-3-031-19833-5_42, ijcai2021p148, yan2023imagespecificinformationsuppressionimplicit} infer region–phrase matches from implicit signals, yet sparse labels leave these alignments weakly constrained and unstable. Fully implicit feature learning \cite{shao2022learninggranularityunifiedrepresentationstexttoimage, shu2022finermoreimplicitmodality, WANG2022108891} optimizes local losses but overlooks the semantics of specific text–region pairs, offering no guarantee of precise correspondences. Masked-modeling–style alignment \cite{irra, fujii2023bilmabidirectionallocalmatchingtextbased} encourages grounding via cross-modal reconstruction, but global-context shortcuts can bypass true dependencies. Across these lines, the core issue persists: locality constraints remain too weak, diffusing supervision, weakening discriminative region selection, and limiting robust local feature learning.

\begin{figure}[t]
    \centering
    \includegraphics[scale=0.5]{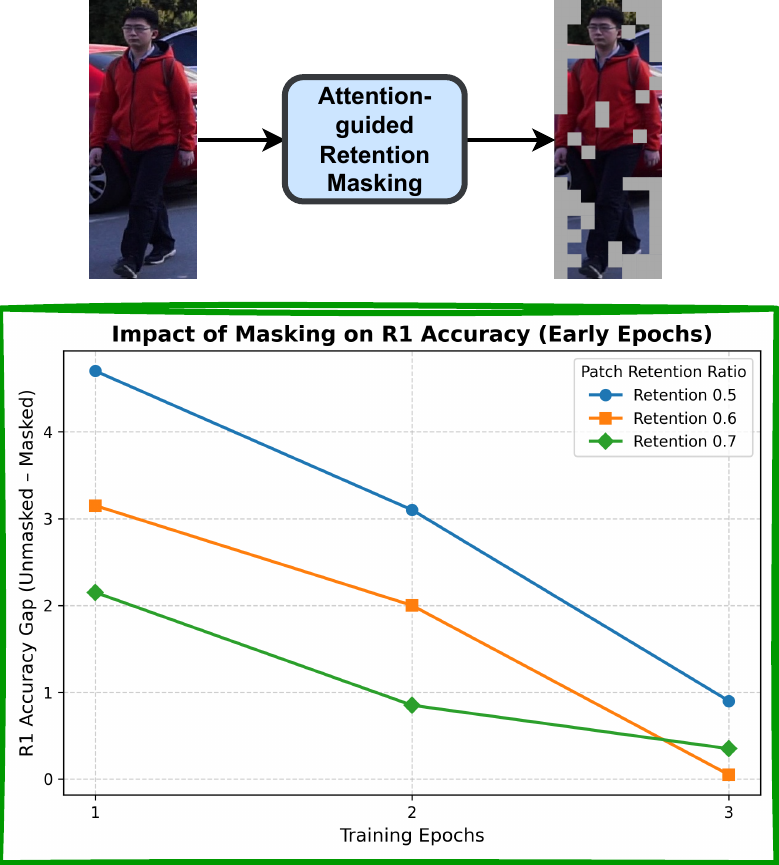}
    \caption{Rank-1 accuracy gap between unmasked and masked images under different mask retention ratios during the early training epochs on RSTP dataset.}
    \label{fig1:observation}
\end{figure}

Attention is well suited to surface fine-grained cues and strengthen cross-modal correspondence, yet its potential remains underexplored thoroughly in TBPS. We probe this with a simple diagnostic: \emph{attention-guided retention masking} (\textbf{\textcolor{red}{\cref{fig1:observation}}}). For each image, we compute attention from the image encoder’s last layer, retain the top-$k$ patches, and mask the rest. We then measure the R1 accuracy gap between the original input and its counterpart over early epochs and multiple retention ratios on RSTP-Reid \cite{zhu2021dssldeepsurroundingspersonseparation}. This analysis reveals two consistent patterns. First, saliency appears early: by epoch 3, the R1 gap falls below 1 percentage point for every retention setting, indicating that the retained patches already capture nearly all discriminative evidence. Second, attention is spatially precise: selected patches consistently align with semantically meaningful parts and carried objects, providing reliable localization cues for discriminative region–phrase correspondence.

Building on these findings, we present \emph{ITSELF}, an attention-guided framework for \emph{implicit local alignment} in TBPS. ITSELF optimizes a global text–image alignment loss and augments it with Guided Representation with Attentive Bank (GRAB), an attention-driven local branch that builds an Attentive Bank by selecting discriminative tokens from both modalities using model’s own attention without additional supervision (\textbf{{\textcolor{red}{\cref{fig1:motivation}(c)}}}). Unlike previous methods that keep local alignment fully implicit, providing no direct constraints on phrase–region correspondence and thus inviting shortcut learning and spurious correlations that cause misalignment, our method injects an attention-derived locality prior that focuses learning on the most informative regions and aligns them consistently across modalities, thereby suppressing noise and reinforcing global alignment. In essence, our key innovation is an attention-driven implicit locality mechanism that turns internal saliency maps into reliable anchors for fine-grained alignment.

Based on analyses \cite{abnar-zuidema-2020-quantifying,voita-etal-2019-analyzing,chefer2021transformerinterpretabilityattentionvisualization} showing different attention heads specialize in complementary cues, with distinct patterns emerging at different depths. We introduce Multi-Layer Attention for Robust Selection (\emph{MARS}) inside \emph{GRAB} to avoid repeatedly picking the same dominant tokens from a single attention map. MARS aggregates attention across layers and performs diversity-aware top-$k$ selection \emph{across modalities}, ensuring complementary coverage. The selected tokens populate GRAB’s Attentive Bank, where local objectives reinforce both inter- and intra-modal structure. 

As suggested by (\textbf{\textcolor{red}{\cref{fig1:observation}}}) where the R1 gap between unmasked and attention-retained inputs quickly narrows in early epochs, we further introduce an \emph{Adaptive Token Scheduler} (ATS) that maintains a larger retention budget at the start to avoid discarding important cues and the resulting training instability, then progressively anneals the budget to focus on high-confidence, fine-grained tokens. This schedule reduces redundancy and false negatives and stabilizes local learning. Finally, following recent practice, we adopt \emph{CLIP} \cite{clip} as the backbone, allowing ITSELF to transfer pretrained knowledge while continuing to learn cross-modal, implicit local correspondences on TBPS. In summary, our main contributions are as follows:

\begin{itemize}
\item \textbf{ITSELF Framework:} A novel attention-guided implicit local alignment framework, \emph{ITSELF}, with \emph{GRAB} leveraging encoder attention to mine fine-grained cues and reinforce global alignment without additional supervision.
\item \textbf{Robust Selection \& Scheduling:} We propose \emph{MARS}, which fuses attention across layers and performs diversity-aware top-$k$ selection; and \emph{ATS}, which anneals the retention budget from coarse to fine over training to stabilize learning and prevent early information loss.
\item \textbf{Strong Empirical Results:} Extensive experiments establish SOTA performance on 3 widely used TBPS benchmarks and improved cross-dataset generalization, confirming the effectiveness and robustness of our approach.
\end{itemize}

\section{Related Work}
\label{sec:related_work}

\subsection{Text-based Person Search (TBPS)}
Over the last few years, the computer vision community has shown a lot of interest in TBPS \cite{irra, rasa, tbpsclip, rde, plot}. With the rise of Vision-Language Pretraining such as \cite{albef,clip}, TBPS research increasingly uses large-scale pretraining to achieve stronger cross-modal representations. A recent line of work enhances TBPS performance by incorporating auxiliary signals. For example, some methods utilize human parsing or pose estimation \cite{sapsam, vitaa} to highlight semantic regions, while others adopt external REID datasets \cite{luperson, dp, unipt, plip, cfam, sapsam} to better adapt to pedestrian domain. These strategies improve fine-grained alignment but introduce additional training cost, annotation dependency, or domain bias. In contrast, our method autonomously extracts and aligns fine-grained local features from both modalities without relying on external datasets or tools, effectively addressing granularity and information gaps in TBPS.

\subsection{Local Alignment for TBPS}
Previous studies enhance fine-grained alignment using either explicit or implicit methods. Explicit approaches leverage external cues, such as human parsing networks \cite{vitaa, sapsam} or large-scale pretraining \cite{aptm, nam}. However, their reliance on extensive external supervision, extra annotations and computational resources often limits their generalization. \noindent In contrast, implicit methods learn local correspondences directly through network without external data. While this removes the dependency on annotations, these methods often suffer from weak semantic grounding, as the relationship between textual descriptions and specific image regions is not explicitly enforced \cite{shu2022finermoreimplicitmodality, cfine, irra}. Consequently, it remains uncertain whether the learned representations truly capture fine-grained cross-modal details. \noindent Our method builds upon implicit, annotation-free work, but introduces a key distinction. Instead of using external models or learning weakly grounded features, we directly exploit the intrinsic attention maps within CLIP. By mining fine-grained cues across multiple layers and selecting the most informative regions, our approach produces more discriminative representations. This lightweight design improves local alignment without requiring additional supervision or pretraining.

\section{Methodology}
\label{sec:Methodology}
This section provides an overview of our proposed framework, ITSELF, in \textbf{\textcolor{red}{\cref{subsec:itself}}}. We then detail the core mechanism, GRAB, which incorporates MARS and ATS, in \textbf{\textcolor{red}{\cref{subsec:grab}}}. Finally, \textbf{\textcolor{red}{\cref{subsec:train}}} presents the training strategy and inference process of the overall pipeline.

\begin{figure*}
    \centering
    \includegraphics[scale=0.5]{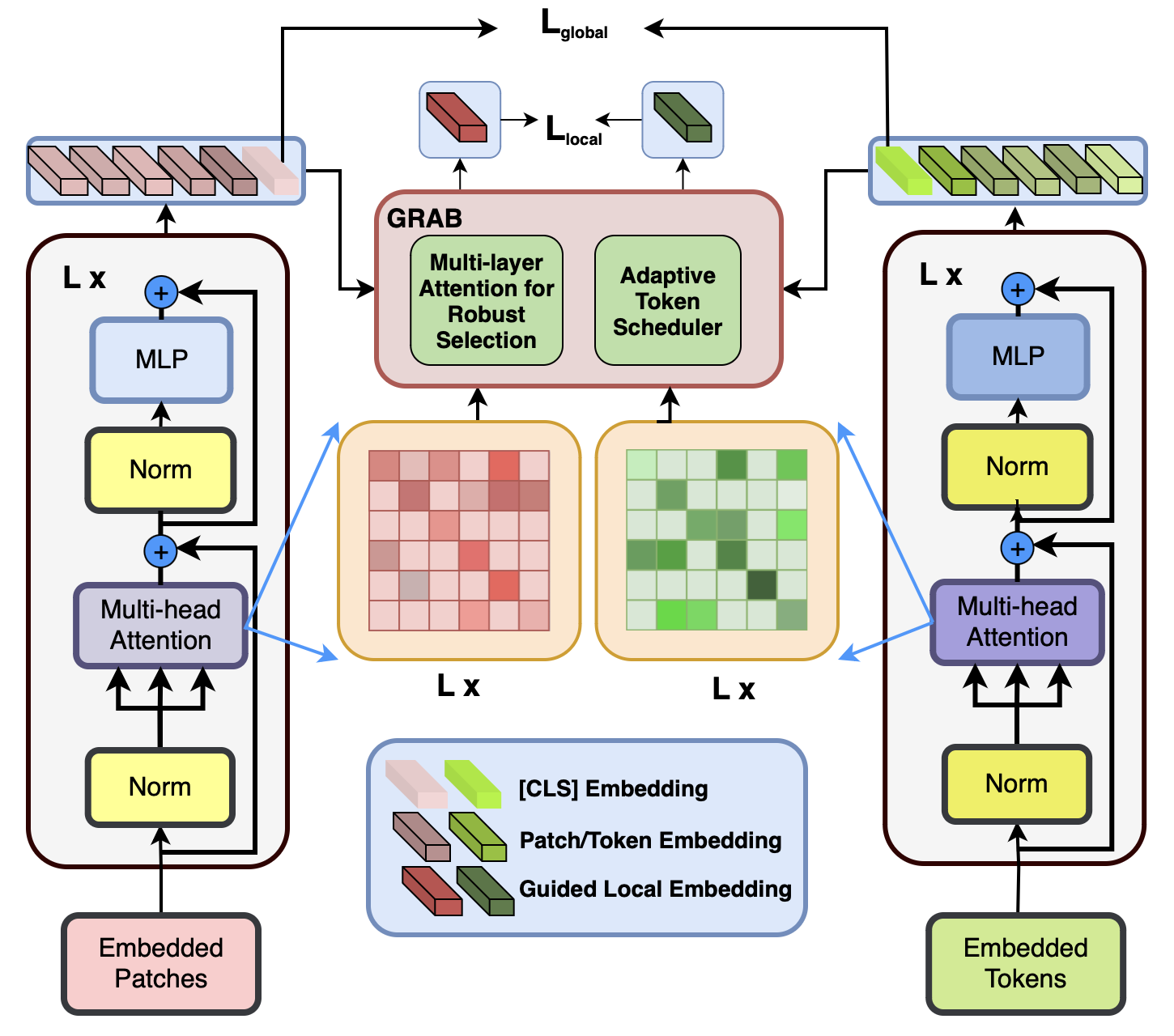}
     \caption{\textbf{Overview of our proposed ITSELF} (an attention-guided implicit local alignment framework). The architecture features a dual-stream encoder for images (left) and text (right). At its core is the GRAB (Guided Representation with Attentive Bank) module, which is designed to learn fine-grained, discriminative cues. GRAB is composed of two key components: MARS (Multi-layer Attention for Robust Selection), which fuses attention across layers to select informative patches/tokens, and ATS (Adaptive Token Scheduler), which anneals the token selection from coarse to fine during training. The model is optimized with a dual-loss strategy: a local loss $L_{local}$ aligns the guided local representations, and a global loss $L_{global}$ matches the final overall embeddings. This allows ITSELF to reinforce global text-image alignment without requiring additional supervision or adding any inference-time cost.}
    \label{fig:overview}
\end{figure*}

\subsection{ITSELF Framework}
\label{subsec:itself}
Our framework consists of three main components \textbf{\textcolor{red}{\cref{fig:overview}}}: (a) an Image Encoder $f_{v}$ that encodes images into embeddings, (b) a Text Encoder $f_{t}$ that generates textual embeddings from captions, and (c) GRAB (Guided Representation with Attentive Bank), which leverages the model’s own attention to construct an attentive bank of high-saliency tokens. Local objectives are then applied on this bank, enabling the model to learn fine-grained correspondences without requiring additional supervision. Following prior works \cite{irra, plot, rde}, we adopt CLIP ViT-B/16 as the backbone for both the visual and textual modalities.

\noindent \textbf{Image Encoder}: Given an input image $I_i \in V$, we divide it into $N=H\times W / P^2$ non-overlapping patches of size $P$, flatten and project them into a $D$-dimensional space, and prepend a learnable [CLS] token with positional embeddings. The sequence is fed into a transformer encoder, yielding visual embeddings $\mathcal{V}_i = f_v(I_i) = \{ v^i_{\text{global}}, v^i_{local} \} \in \mathbb{R}^{(1+N)\times D}$, where ${v^i_{global}} = {v}^i_{\text{cls}}$ is the global embedding and $v^i_{local} = \{v^i_j\}_{j=1}^N$ are patch embeddings.

\noindent \textbf{Text Encoder}: For text, we adopt CLIP’s Transformer-based encoder. Given a caption $T_i \in T$, it is tokenized with BPE and wrapped with [SOS]/[EOS] tokens. The sequence is embedded and passed through the transformer to produce
$\mathcal{T}_i = f_t(T_i) = \{ t^i_{\text{global}}, t^i_{\text{local}} \} \in \mathbb{R}^{(L+2)\times D},$
where $t^i_{\text{global}} = t^i_e$ (from [EOS]) is the global embedding, and $t^i_{\text{local}} = \{t^i_j\}_{j=1}^L$ are token-level embeddings. The [SOS] token $t^i_s$ is retained but unused.

\subsection{Guided Representation with Attentive Bank}
\label{subsec:grab}

\subsubsection{Multi-layer Attention for Robust Selection}


To learn robust implicit local representations, we design the GRAB, which retains a diverse set of highly discriminative tokens. Building on our finding \textbf{\textcolor{red}{\cref{fig1:observation}}} that tokens with consistently high attention values encode core identity cues, even from the earliest training epochs. However, selecting tokens based solely on a single fixed layer is inherently suboptimal, since different Transformer layers capture different types of information: shallow layers emphasize low-level textures, middle layers capture broader context, and deeper layers encode semantic abstractions that may suppress fine-grained details. To overcome this limitation, we introduce \textbf{Multi-layer Attention for Robust Selection (MARS)}, which aggregates attention information across multiple layers to provide a more stable and reliable estimate of patch importance. Formally, given attention maps \(\mathbf{A}^{(\ell)} \in \mathbb{R}^{  N \times N}\) from selected layers \(\ell \in \mathcal{L}\), we denoise by removing the lowest \(\delta_\ell\) fraction of attention weights, thereby filtering out non-informative links. The pruned attention map is then combined with the identity matrix \textbf{I} to preserve self-dependencies and normalized as:
\begin{equation}
\hat{\mathbf{A}}^{(\ell)} = \operatorname{Norm}\left(\frac{\operatorname{Discard}\left(\mathbf{A}^{(\ell)}, \delta_\ell\right) + \mathbf{I}}{2}\right),
\end{equation}

where, \(\operatorname{Discard}(\cdot, \delta_\ell)\) sets the lowest \(\delta_\ell\) proportion of elements to zero while preserving the rest.  

The aggregated attention across the selected layers is then computed by sequential composition:
\begin{equation}
\mathbf{R} = \prod_{\ell \in \mathcal{L}} \hat{\mathbf{A}}^{(\ell)}_b, 
\quad \mathbf{R} \in \mathbb{R}^{ N \times N}.
\end{equation}

Based on the aggregated map $\mathbf{R}$, we obtain local features $\mathbf{f}^{\text{loc}}_{i}, \ \mathbf{f}^{\text{loc}}_{t}$  by selecting the most informative tokens:
\begin{equation}
\mathbf{f}^{\text{loc}}_{i}, \ \mathbf{f}^{\text{loc}}_{t} 
= \operatorname{TopK}\big(\mathbf{R}_{i}, \mathbf{R}_{t}, v^i_{local}, t^i_{local}),
\end{equation}
where $\mathbf{R}_{i}$ and $\mathbf{R}_{t}$ denote the aggregated attention maps for image and text modalities. 


By ranking and selecting the diversity-aware top-$k$ tokens according to $\mathbf{R}$, MARS implicitly captures fine-grained feature, amplify discriminative signals, suppress background noise, strengthens the global embedding and yields a more robust, identity-aware feature space.

\subsubsection{Adaptive Token Scheduler}



To further strengthen local feature learning, we introduce the \emph{Adaptive Token Scheduler} (ATS). Unlike a fixed token budget, which may retain excessive background information or prematurely discard useful details, ATS \emph{anneals} the number of selected tokens over training, This design follows a coarse-to-fine paradigm: in early stages, a larger proportion of tokens is preserved to avoid losing critical identity cues; in later stages, the selection increasingly focuses on highly discriminative patches. Formally, the number of retained tokens at step $t$ is defined as:

\begin{equation}
k_t =
\begin{cases}
\left\lfloor N\, \rho_{\text{start}} \left(\dfrac{\rho_{\text{end}}}{\rho_{\text{start}}}\right)^{\tfrac{t}{T}} \right\rfloor, & \text{if } t \le T, \\[8pt]
\left\lfloor N\, \rho_{\text{end}} \right\rfloor, & \text{if } t > T,
\end{cases}
\end{equation}
where $N$ is the number of tokens, $\rho_{\text{start}}$ and $\rho_{\text{end}}$ are the initial and final retention ratios, $t$ is the current training step, and $T$ is the schedule length. This gradual narrowing mitigates early information loss and stabilizes training, while ultimately emphasizing fine-grained signals that complement the global embedding.

\subsubsection{Implicit Local Learning Alignment}

\noindent\textbf{Algorithm description.} 
To populate the \textit{Attentive Bank} with the most discriminative cues from both modalities, we score each patch/token by the aggregated attention $\mathbf{R}$ returned by \textsc{MARS}. In early training, to stabilize optimization and avoid missing salient regions, we use \textsc{ATS} to schedule a decaying token budget
$k$ decreasing over steps. At each step, we select the top-$k$ patches/tokens with the highest attention in $\mathbf{R}$ and insert them into the bank. To \emph{bridge modality shift and refine local features while preserving the original signal}, the selected tokens are passed through a lightweight \emph{Adapter} (MLP) with a residual connection. We then apply $\mathrm{GPO}$ \cite{chen2021learningbestpoolingstrategy} to obtain Guided Local Embedding $\mathbf{v}, \boldsymbol{\tau}$ and compute the local-alignment loss $\mathcal{L}_{\text{local}}$. The complete procedure is summarized in \textbf{\textcolor{red}{\cref{alg:grab}}}.

\begin{algorithm}[t]
\caption{GRAB: Implicit Local Alignment}
\label{alg:grab}
\begin{algorithmic}[1]
\Require Patch/Token features \( v^i_{local}, v^t_{local}\);
         attention maps \(\{\mathbf{A}^{(\ell)}\}\); 
          \(\text{Adapter}(\cdot)\), \(\GPO(\cdot)\); loss \(\mathcal{L}_{\text{local}}(\cdot,\cdot)\)
\Ensure Local-alignment loss \(\mathcal{L}\)
\State \textbf{Aggregated attention (MARS):} \(\mathbf{R}_i, \mathbf{R}_t\leftarrow\) calculated by Eq.~(2).
\State \textbf{Token budget (ATS):} \(k \leftarrow\) calculated by Eq.~(4).
\State \textbf{Select informative tokens:} 
       \((\mathbf{f}^{\mathrm{loc}}_i,\mathbf{f}^{\mathrm{loc}}_t) \leftarrow\) calculated by Eq.~(3) using \(\mathbf{R}_i,\mathbf{R}_t\) and \(k\).

\State \textbf{Local adaptation (residual):}
       \((\mathbf{f}^{\mathrm{loc}}_i,\mathbf{f}^{\mathrm{loc}}_t)
       \leftarrow \text{Adapter}(\mathbf{f}^{\mathrm{loc}}_i,\mathbf{f}^{\mathrm{loc}}_t) + (\mathbf{f}^{loc}_i,\mathbf{f}^{loc}_t)\).
\State \textbf{Pooling:}
       \(\mathbf{v} \leftarrow \GPO(\mathbf{f}^{\mathrm{loc}}_i)\), \quad
       \(\boldsymbol{\tau} \leftarrow \GPO(\mathbf{f}^{\mathrm{loc}}_t)\).
\State \textbf{Loss:} \(\mathcal{L} \leftarrow \mathcal{L}_{\text{local}}(\mathbf{v}, \boldsymbol{\tau})\).
\State \Return \(\mathcal{L}\).
\end{algorithmic}
\end{algorithm}

\subsection{Training and Inference}
\label{subsec:train}
\textbf{Training.} 
To optimize both global appearance and fine-grained discriminative tokens, our approach combines Triplet Alignment Loss (TAL) \cite{rde} and Cross-Modal Identity Loss (CID) \cite{cfam}. By applying these specialized losses to separate global and local embeddings, the model learns a more comprehensive and robust representation for matching textual descriptions to pedestrian images.

\noindent Given an image-text pair $(I,T)$, the loss is defined as:

\begin{equation}
\mathcal{L} = \mathcal{L}_{tal} + \mathcal{L}_{cid}
\end{equation}

\noindent We apply this loss to the global embedding $(v_{global}, t_{global})$ and the guided local representation $(\mathbf{v}, \tau)$, yielding
\begin{equation}
    \mathcal{L}_{global} = \mathcal{L}(v_{global}, t_{global}), \mathcal{L}_{local} = \mathcal{L}(\mathbf{v}, \tau)
\end{equation}
\noindent The overall training objective is the sum of the two:
\begin{equation}
    \mathcal{L}_{total} = \mathcal{L}_{global} + \mathcal{L}_{local}
\end{equation}
\noindent \textbf{Inference.} In the inference process, the final image-text pair similarity is computed by combining both global and local similarity. Here is the specific computation formula:
\begin{equation}
    \mathcal{S} = \lambda_S\times\mathcal{S}_{global} + (1-\lambda_S)\times\mathcal{S}_{local}
\end{equation}
where $\mathcal{S}_{global}$ and $\mathcal{S}_{local}$ represent global similarity and local similarity, respectively, and $\lambda_S$ is the weighting factor.

\section{Experiment}
\label{sec:experiment}
\subsection{Experimental Setup}
\textbf{Datasets.} We evaluate the effectiveness of the proposed method on widely used public datasets for TBPR tasks, including CUHK-PEDES \cite{cuhk}, ICFG-PEDES \cite{icfgpedes}, and RSTPReid \cite{rstpreid}. Additional details about these datasets are provided in the Supplementary Material.

\noindent \textbf{Evaluation Metrics.} For evaluation, we employ the widely used Rank-k accuracy (k = 1, 5, 10) and mean Average Precision (mAP) metrics across all three datasets.

\subsection{Quantitative Results}

\begin{table*}[htp]
\centering
\resizebox{\textwidth}{!}{
\begin{tabular}{c c c c | c c c c | c c c c | c c c c}
\toprule
\multirow{2}{*}{Method} & \multirow{2}{*}{Venue} & \multirow{2}{*}{Image enc.} & \multirow{2}{*}{Text enc.} & 
\multicolumn{4}{c|}{CUHK-PEDES} & 
\multicolumn{4}{c|}{ICFG-PEDES} & 
\multicolumn{4}{c}{RSTP-Reid} \\
\cmidrule(lr){5-8} \cmidrule(lr){9-12} \cmidrule(lr){13-16}
 & & & & R@1 & R@5 & R@10 & mAP & R@1 & R@5 & R@10 & mAP & R@1 & R@5 & R@10 & mAP \\

\midrule
 \multicolumn{16}{l}{with ALBEF backbone:} \\
\midrule
RaSa \cite{rasa} & IJCAI'23 & Swin-B & BERT &
76.51 & 90.29 & 94.25 & 69.38 &
65.28 & 80.40 & 85.12 & 41.29 &
65.20 & 84.05 & 89.85 & 50.14 \\
MARS \cite{mars} & TOMM'25 & Swin-B & BERT &
77.62 & 90.63 & 94.27 & 71.71 &
67.60 & 81.47 & 85.79 & 44.93 &
67.55 & 86.65 & 91.35 & 52.92 \\
\midrule
\multicolumn{16}{l}{with using external tools or ReID-Domain Pre-training:} \\
\midrule
UniPT \cite{shao2023unifiedpretrainingpseudotexts} & ICCV'23 & ViT-B/16 & BERT & 68.50 & 84.67 & 90.38 & - &
60.09 & 76.19 & 82.46 & - &
51.85 & 74.85 & 82.85 & - \\
CFAM (B/16) \cite{cfam}& CVPR’24 & CLIP-ViT & CLIP-Xformer & 72.87 & 88.61 & 92.87 & 64.92 & 62.17 & 79.57 & 85.32 & 36.34 & 59.40 & 81.35 & 88.50 & 46.04 \\
SAP-SAM \cite{sapsam}& MM’24 & CLIP-ViT & CLIP-Xformer & 75.05 & 89.93 & 93.73 & - & 63.97 & 80.84 & 86.17 & - & 62.85 & 82.65 & 89.85 & - \\
CFAM(L/14) \cite{cfam} & CVPR'24 & CLIP-ViT & CLIP-Xformer & 75.60 & 90.53 & 94.36 & 67.27 & 65.38 & 81.17 & 86.35 & 39.42 & 62.45 & 83.55 & 91.10 & 49.50 \\
DP \cite{dp}& AAAI'24 & CLIP-ViT & CLIP-Xformer & 75.66 & 90.59 & 94.07 & 66.58 & 65.61 & 81.73 & 86.95 & 39.14 & 62.48 & 83.77 & 89.93 & 48.86 \\
APTM \cite{aptm} & MM'23 & Swin-B & BERT &
76.53 & 90.04 & 94.15 & 66.91 &
68.51 & 82.99 & 87.56 & 41.22 &
67.50 & 85.70 & 91.45 & 52.56 \\
AUL \cite{aul}& AAAI'24 & Swin-B & BERT &
77.23 & 90.43 & 94.41 & - &
69.16 & 83.32 & 88.37 & - &
71.65 & 87.55 & 92.05 & - \\
CAMeL \cite{camel} & TIFS'25 & SG-Former & BERT &
77.24 & 91.80 & 95.16 & 68.32 &
68.70 & 83.11 & 88.32 & 41.58 &
68.50 & 87.40 & 92.70 & 53.61 \\
\midrule
\multicolumn{16}{l}{with CLIP backbone:} \\
\midrule
CLIP (ViT-B/16) \cite{clip} & ICML'21 & CLIP-ViT & CLIP-Xformer & 65.73 & 86.39 & 92.01 & 61.97 & 56.23 & 74.29 & 81.62 & 30.85 & 56.67 & 78.09 & 86.62 & 42.85 \\
CFine \cite{cfine} & TIP’23 & CLIP-ViT & BERT & 69.57 & 85.93 & 91.15 & - & 60.83 & 76.55 & 82.42 & - & 50.55 & 72.50 & 81.60 & - \\
CSKT \cite{cskt} & ICASSP'24 & CLIP-ViT & CLIP-Xformer & 69.70 & 86.92 & 91.80 & 62.74 & 58.90 & 77.31 & 83.56 & 33.87 & 57.75 & 81.30 & 88.35 & 46.43 \\
DM-Adapter \cite{dmadapter} & AAAI’25 & CLIP-ViT & CLIP-Xformer & 72.17 & 88.74 & 92.85 & 64.33 & 62.64 & 79.53 & 85.32 & 36.50 & 60.00 & 82.10 & 87.90 & 47.37 \\
IRRA \cite{irra}& CVPR’23 & CLIP-ViT & CLIP-Xformer & 73.38 & 89.93 & 93.71 & 66.13 & 63.46 & 80.25 & 85.82 & 38.06 & 60.20 & 81.30 & 88.20 & 47.17 \\
TBPSCLIP \cite{tbpsclip}& AAAI’24 & CLIP-ViT & CLIP-Xformer & 73.54 & 88.19 & 92.35 & 65.38 & 65.05 & 80.34 & 85.47 & 39.83 & 61.95 & 83.55 & 88.75 & 48.26 \\
BiLMa \cite{bilma} & ICCV'23 & CLIP-ViT & CLIP-Xformer & 74.03 & 89.59 & 93.62 & 66.57 & 63.83 & 80.15 & 85.74 & 38.26 & 61.20 & 81.50 & 88.80 & 48.51 \\
MUM \cite{mum}& AAAI’24 & CLIP-ViT & CLIP-Xformer & 74.25 & 89.83 & 93.58 & 66.15 & 65.62 & 80.54 & 85.83 & 38.78 & 63.40 & 83.30 & \textcolor{blue}{90.30} & 49.28 \\
CRUE \cite{crue} & TOMM'25 & CLIP-ViT & CLIP-Xformer & 74.91 & 90.11 & 93.90 & \textcolor{blue}{68.93} & 64.88 & 80.90 & 86.36 & \textcolor{blue}{41.82} & 63.15 & 82.80 & 90.20 & 50.20 \\
PLOT \cite{plot}& ECCV’24 & CLIP-ViT & CLIP-Xformer & 75.28 & \textcolor{blue}{90.42} & \textcolor{blue}{94.12} & - & 65.76 & 81.39 & 86.73 & - & 61.80 & 82.85 & 89.45 & - \\
RDE \cite{rde} & CVPR’24 & CLIP-ViT & CLIP-Xformer & \textcolor{blue}{75.94} & 
90.14 & 
\textcolor{blue}{94.12} & 
67.56 & 
\textcolor{blue}{67.68} & 
\textcolor{blue}{82.47} & 
\textcolor{blue}{87.36} & 
40.06 & 
\textcolor{blue}{65.35} & 
\textcolor{blue}{83.95} & 
89.90 & 
\textcolor{blue}{50.88} \\
\midrule
\textbf{Ours} & \textbf{WACV'26} & CLIP-ViT & CLIP-Xformer & \textbf{\textcolor{red}{76.95}} & \textbf{\textcolor{red}{90.64}} & \textbf{\textcolor{red}{94.36}} & \textbf{\textcolor{red}{69.38}} & 
\textbf{\textcolor{red}{69.23}} & 
\textbf{\textcolor{red}{82.84}} & 
\textbf{\textcolor{red}{87.62}} & 
\textbf{\textcolor{red}{43.80}} & 
\textbf{\textcolor{red}{67.30}} & 
\textbf{\textcolor{red}{85.60}} & 
\textbf{\textcolor{red}{90.50}} & 
\textbf{\textcolor{red}{53.05}} \\

\bottomrule
\end{tabular}}
\caption{Performance of text-based person search methods on three datasets.}
\label{tab:all_datasets}
\end{table*}


\textbf{Comparison with state-of-the-art methods.} We present comparison results with SOTA methods on three widely used benchmark datasets \textbf{\textcolor{red}{\cref{tab:all_datasets}}}. Our approach clearly stands out, outperforming all CLIP-based competitors on every metric. Within CLIP-backbone methods, we set new SOTA \textbf{R@1} on all datasets, improving over the strongest prior by \textbf{+1.01}\%, \textbf{+1.55}\%, and \textbf{+1.95}\%, and we also obtain the best \textbf{mAP} on every benchmark, with the largest gain on \textbf{RSTP-Reid} (\textbf{+2.17}\% mAP). These gains persist at deeper ranks, \textbf{R@5}/\textbf{R@10} are best or tied-best, indicating broad retrieval improvements rather than a single-metric bump. Crucially, we achieve this \emph{without} ReID-domain pretraining by mining multi-layer attention (MARS) to guide implicit local alignment. Despite this minimalist setup, we still surpass methods that leverage extra resources or larger backbones (e.g., CFAM(L/14), DP, UniPT). Notably, using only CLIP, we attain the top \textbf{R@1} on \textbf{ICFG-PEDES} over all methods.


\noindent\textbf{Domain Generalization.} We further evaluate the cross-domain robustness of our model by training on one source dataset and directly testing on another unseen target dataset without fine-tuning. Using CUHK-PEDES (C), ICFG-PEDES (I), and RSTPReid (R), we form six transfer settings (e.g., C→I, I→R). Existing local implicit matching methods such as IRRA often achieve strong in-domain performance but generalize poorly, mainly because their relational modules overfit to dataset-specific patterns such as clothing styles or annotation bias. As a result, their learned correspondences fail to transfer effectively across domains. In contrast, our method avoids such overfitting by integrating attentive bank guidance with balanced local–global alignment, enabling more semantically stable representations. As shown in \textbf{\textcolor{red}{\cref{tab:cross}}}, our approach achieves the best results in all six transfer settings. For instance, in the C→I scenario, we obtain 50.58\% R@1 and 27.32\% mAP, surpassing RDE by \textbf{2.40\%} and \textbf{2.32\%}. Similarly, in the challenging I→R transfer, we improve the previous best R@1 by over 2\%. These results demonstrate that unlike prior local implicit methods, our model achieves both high performance and strong domain generalization—crucial for real-world TBPS scenarios with diverse and shifting data distributions.
\begin{table}[htp]
\begin{tabular}{c|lcccc}
\toprule
      & Method & R1 & R5 & R10 & mAP \\ 
\midrule
C$\to$I & IRRA \cite{irra} & 42.41 & 62.11 & 69.62 & 21.77 \\ 
        & CLIP \cite{clip} & 43.04 & - & - & 22.45 \\ 
        & RDE \cite{rde} & 48.18 & 66.30 & 73.70 & 25.00 \\ \cline{2-6}
        & \textbf{Ours} & \textbf{50.58} & \textbf{67.81} & \textbf{74.68} & \textbf{27.32} \\
\midrule
I$\to$C & IRRA \cite{irra} & 33.48 & 56.29 & 66.33 & 31.56 \\ 
        & CLIP \cite{clip} & 33.90 & - & - & 31.65 \\ 
        & RDE \cite{rde} & 38.11 & 59.24 & 68.44 & 34.16 \\ \cline{2-6}
        & \textbf{Ours} & \textbf{41.05} & \textbf{63.30} & \textbf{72.08} & \textbf{37.79} \\
\midrule
I$\to$R & IRRA \cite{irra} & 45.30 & 69.25 & 78.80 & 36.82 \\ 
        & CLIP \cite{clip} & 47.45 & - & - & 36.83 \\ 
        & RDE \cite{rde} & 49.25 & 72.10 & 80.20 & 38.46 \\ \cline{2-6}
        & \textbf{Ours} & \textbf{51.30} & \textbf{73.30} & \textbf{80.40} & \textbf{41.07} \\ 
\midrule
R$\to$I & IRRA \cite{irra} & 32.30 & 49.67 & 57.80 & 20.54 \\ 
        & CLIP \cite{clip} & 33.58 & - & - & 19.58 \\ 
        & RDE \cite{rde} & 42.17 & 58.32 & 65.49 & 26.37 \\ \cline{2-6}
        & \textbf{Ours} & \textbf{43.32} & \textbf{59.17} & \textbf{66.16} & \textbf{27.57} \\ 
\midrule
C$\to$R & CLIP \cite{clip} & 52.55 & - & - & 39.97 \\ 
        & IRRA \cite{irra} & 53.25 & 77.15 & 85.35 & 39.63 \\ 
        & RDE \cite{rde} & 54.90 & 77.50 & 86.50 & 41.27 \\ \cline{2-6}
        & \textbf{Ours} & \textbf{58.05} & \textbf{79.30} & \textbf{86.85} & \textbf{43.72} \\ 
\midrule
R$\to$C & IRRA \cite{irra} & 32.80 & 55.26 & 65.81 & 30.29 \\ 
        & CLIP \cite{clip} & 35.25 & - & - & 32.35 \\ 
        & RDE \cite{rde} & 36.94 & 58.22 & 67.58 & 33.65 \\ \cline{2-6}
        & \textbf{Ours} & \textbf{40.32} & \textbf{61.70} & \textbf{71.04} & \textbf{36.65} \\ 
\bottomrule
\end{tabular}
\caption{Comparisons with state-of-the-arts(domain generalization). Here “C” denotes CUHK-PEDES, “I” represents ICFG-PEDES and "R" means RSTPReid.}
\label{tab:cross}
\end{table}

\begin{table*}[htp]
\centering
\resizebox{\textwidth}{!}{
\begin{tabular}{c|ccc|cccc|cccc|cccc}
\hline
\multirow{2}{*}{Method} & \multicolumn{3}{c|}{GRAB} & \multicolumn{4}{c|}{CUHK-PEDES} & \multicolumn{4}{c|}{ICFG-PEDES} & \multicolumn{4}{c}{RSTPReid} \\
\cmidrule(lr){5-8} \cmidrule(lr){9-12} \cmidrule(lr){13-16}
 & SL & MARS & ATS & R@1 & R@5 & R@10 & mAP & R@1 & R@5 & R@10 & mAP & R@1 & R@5 & R@10 & mAP \\
\hline
Baseline & \textcolor{red}{\ding{55}} & \textcolor{red}{\ding{55}} & \textcolor{red}{\ding{55}} & 
74.66 & 89.70 & 93.57 & 67.63 & 
66.06 & 81.12 & 86.10 & 41.02 & 
61.30 & 80.85 & 87.65 & 49.12 \\
+SL       & \textcolor{green}{\ding{51}} & \textcolor{red}{\ding{55}} & \textcolor{red}{\ding{55}} & 
76.49 & 90.20 & 94.35 & 58.97 & 
67.93 & 82.36 & 87.12 & 42.38 & 
65.00 & 85.45 & 90.15 & 52.26 \\
+MARS        & \textcolor{red}{\ding{55}} & \textcolor{green}{\ding{51}} & \textcolor{red}{\ding{55}} & 
76.90 & 90.46 & 94.35 & \textbf{69.71} & 
69.07 & 82.72 & 87.51 & 43.67 & 
66.95 & 85.15 & 90.40 & \textbf{53.05} \\
+SL+ATS        & \textcolor{green}{\ding{51}} & \textcolor{red}{\ding{55}} & \textcolor{green}{\ding{51}} & 76.66 & 90.58 & 94.22 & 69.18 & 68.76 & 82.74 & 87.45 & 43.35 & 65.95 & \textbf{85.60} & 90.10 & 52.71 \\
\midrule
\textbf{Ours}     & \textcolor{red}{\ding{55}} & \textcolor{green}{\ding{51}} & \textcolor{green}{\ding{51}} & 
\textbf{76.95} & \textbf{90.64} & \textbf{94.36} & 69.38 & \textbf{69.23} & \textbf{82.84} & \textbf{87.62} & \textbf{43.80} & \textbf{67.30} & \textbf{85.60} & \textbf{90.50} & \textbf{53.05} \\
$\Delta$ & - & - & - & \textbf{\textcolor{green}{+2.29}} & 
\textbf{\textcolor{green}{+0.94}} & 
\textbf{\textcolor{green}{+0.79}} & 
\textbf{\textcolor{green}{+1.75}} & 
\textbf{\textcolor{green}{+3.17}} & 
\textbf{\textcolor{green}{+1.72}} & 
\textbf{\textcolor{green}{+1.52}} & 
\textbf{\textcolor{green}{+2.78}} & 
\textbf{\textcolor{green}{+6.00}} & 
\textbf{\textcolor{green}{+4.75}} & 
\textbf{\textcolor{green}{+2.85}} & 
\textbf{\textcolor{green}{+3.93}} \\
\hline
\end{tabular}
}
\caption{Ablation study on each component of ITSELF on three datasets.}
\label{tab:benchmarks}
\end{table*}

\subsection{Qualitative Results}
\begin{figure}[htp]
    \centering
    \includegraphics[scale=0.32]{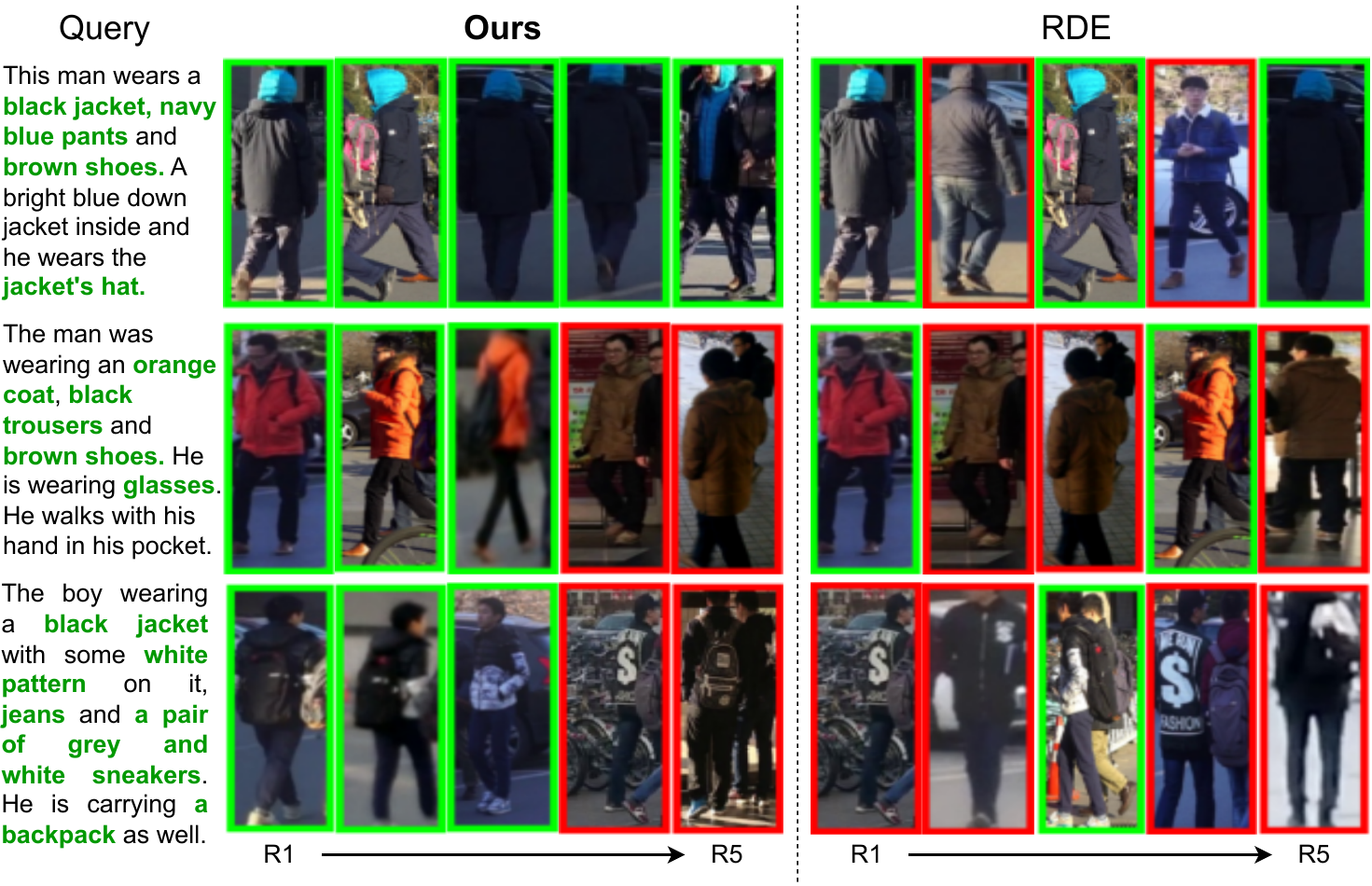}
    \caption{Qualitative results of text-to-image retrieval on RSTPReid benchmark, comparing our method with RDE \cite{rde}. Retrieved images are ranked from left to right in descending order of similarity. Correct matches are outlined in green, while incorrect ones are shown in red. Text highlighted in green indicates the descriptive details effectively captured by our approach.}
    \label{fig:2}
\end{figure}

\noindent \textbf{Top-5 Retrieval Examples.} 
\textbf{\textcolor{red}{\cref{fig:2}}} qualitatively demonstrates our method's superiority over RDE on the RSTPReid benchmark. For a query about a man in a black jacket and blue down jacket, our method retrieves all five correct matches, whereas RDE finds only three. With a query for a man in an orange coat with a hand-in-pocket pose, our approach correctly retrieves three images in the top ranks, while RDE secures only two, ranking the second true positive fourth. For a boy in a black jacket with white patterns, our method again excels with three correct top-ranked retrievals, contrasting with RDE's single correct match at R3. These examples highlight our framework's enhanced ability to align fine-grained textual descriptions with relevant visual details.

\begin{figure}[htp]
    \centering
    \includegraphics[scale=0.55]{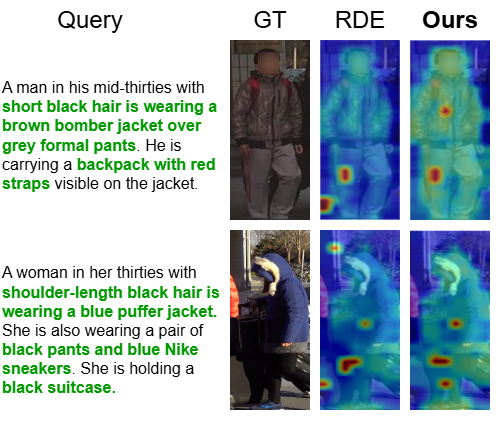}
    \caption{Qualitative comparison of attention maps generated by RDE \cite{rde} and by our method using the Grad-CAM.\cite{gradcam}}
    \label{fig:gradcamdif}
\end{figure}

\noindent \textbf{Attention Comparison.} \textbf{\textcolor{red}{\cref{fig:gradcamdif}}} presents Grad-CAM \cite{gradcam} visualizations comparing RDE \cite{rde} and our method. 
Our approach achieves sharper localization by focusing on query-specific elements (e.g., clothing and accessories) with minimal spillover, effectively isolating the target pedestrian in multi-person scenes.  In contrast, RDE produces diffuse, often irrelevant hotspots, leading to fragmented text-image associations. These results highlight our framework’s stronger attribute-level fidelity and reduced cross-identity confusion, crucial for real-world person search.

\noindent \textbf{Analysis on Top-K token Selection.} The comparison in \textbf{\textcolor{red}{\cref{fig:attn}}} shows our proposed top-K token selection is more effective than the baseline. While the baseline produces a scattered attention map misaligned with the query, our method generates a focused, semantically relevant map. It successfully pinpoints image regions corresponding to keywords demonstrating a superior ability to ground textual descriptions in their correct visual context for accurate retrieval.


\begin{figure}[htp]
    \centering
    \includegraphics[scale=0.4]{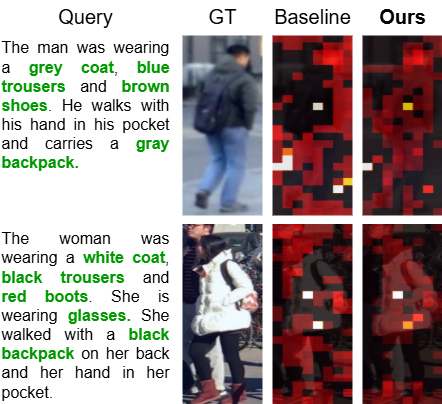}
    \caption{Comparison of selected top-K patches in text-to-image retrieval by the baseline and by our method. Our approach highlights more semantically relevant regions (e.g., clothing colors and accessories) corresponding to the query descriptions.}
    \label{fig:attn}
\end{figure}


\subsection{Ablation Study}

\begin{figure*}[htp]
    \centering
    \includegraphics[width=\textwidth]{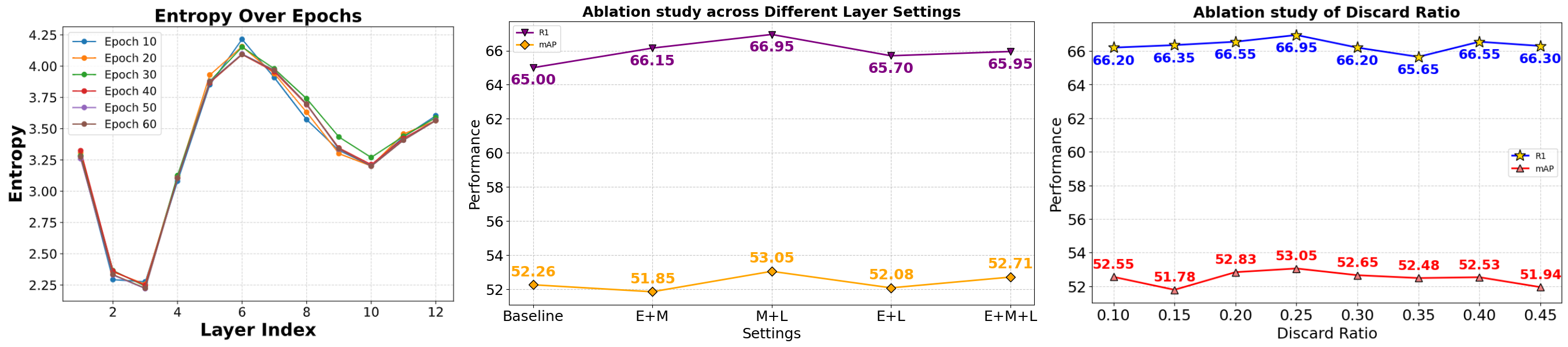}
    \caption{\textbf{(Left)} Entropy of attention per CLIP layer across training epochs. \textbf{(Middle)} Ablation of which CLIP layers are selected for MARS (E: Early, M: Middle, L: Late). \textbf{(Right)} Sensitivity of the MARS discard ratio, evaluated on R\@1 and mAP.}

    \label{fig:huynt}
\end{figure*}

\textbf{Effectiveness of each component:} To comprehensively evaluate the contribution of each component in our proposed ITSELF framework, we conduct a systematic empirical analysis on three publicly available datasets. The detailed experimental results are presented in \textbf{\textcolor{red}{\cref{tab:benchmarks}}}. \textbf{MARS module:} To evaluate the effectiveness of MARS, we first compare it with a fixed single-layer strategy (SL). Following our entropy of attention distributions (\textbf{\textcolor{red}{\cref{fig:huynt} (Left)}}), we select layer 3 as SL since it exhibits the lowest entropy, indicating higher confidence. While both SL and MARS outperform the baseline, MARS consistently achieves superior results across all datasets, with notable R\@1 gains of +2.24\%, +3.01\%, and +5.65\%, demonstrating its clear advantage over relying on a fixed SL. \textbf{ATS Module:} Adding ATS on top of the SL-only baseline already yields consistent gains in R\@1. When combined with MARS in the full model, we achieve the strongest performance overall, with substantial improvements over the baseline (+2.29\%, +3.17\%, +6.00\%) on three datasets. These results confirm that ATS not only prevents the loss of discriminative cues in early training but also contributes to more stable optimization.


\noindent \textbf{Analysis of Layer Selection and Discard Ratio in MARS:} From the middle plot in \textbf{\textcolor{red}{\cref{fig:huynt}}}, multi-layer configurations in MARS consistently outperform the baseline, with the \emph{Middle+Late (M+L)} combination achieving the best R\@1 and mAP among all layer-type combinations (E, M, L and their pairings). The left plot explains why: early layer's attention (notably layer 3) have the lowest entropy, meaning attention is sharply peaked on a few low-level tokens (edges/textures or background), which offers weak semantic grounding and can inject noise when fused. By contrast, middle layers show the higher entropy, capturing broader context and relations, while late layers re-focus attention onto salient, semantic regions. Fusing "M+L" therefore balances contextual coverage with discriminative focus, outperforming any option that includes Early Layer's Attention. Finally, the right plot shows the best performance at a discard ratio of 0.25, indicating that removing a small fraction of low-attention tokens during fusion filters noise and sharpens discriminative cues.

\section{Conclusion}
\label{sec:conclusion
}

In this paper, we introduce \emph{ITSELF}, a novel attention-guided framework for \emph{implicit local alignment} in TBPS that turns CLIP’s multi-layer attention into an Attentive Bank without extra supervision or inference-time cost. Building on this idea, \emph{GRAB} harvests and optimizes fine-grained correspondences through an implicit local objective that complements the global loss. To realize GRAB, \emph{MARS} adaptively aggregates and ranks attention across layers to select the most discriminative patches/tokens. In parallel, \emph{ATS} schedules the token-retention budget from coarse to fine, mitigating early information loss and stabilizing training. Extensive experiments on three widely used TBPS benchmarks demonstrate \textbf{state-of-the-art} performance among CLIP-based methods across all metrics and improved cross-dataset generalization, confirming the effectiveness, robustness, and practicality of our approach.


\section*{Acknowledgement}

This research is supported by VNUHCM-University of Information Technology's Scientific Research Support Fund.

\appendix
\begin{center}
\section*{Supplementary Material}
\addcontentsline{toc}{section}{Supplementary Material}

\end{center}

\section{Experimental Details}
\label{sec:intro}

\textbf{Datasets.} We conduct experiments on three widely used text-to-image person retrieval benchmarks.

\begin{enumerate}
    \item CUHK-PEDES \cite{cuhk} provides 40,206 pedestrian images paired with 80,412 textual descriptions corresponding to 13,003 identities, with splits of 11,003 for training, 1,000 for validation, and 1,000 for testing.

    \item   ICFG-PEDES \cite{icfgpedes} consists of 54,522 image-text pairs from 4,102 individual IDs, which are split into 34,674 and 19,848 for training and testing, respectively.

    \item RSTPReid \cite{rstpreid} contains 20,505 images of 4,101 individual IDs, with each ID having 5 images and each image associated with the corresponding two annotated text descriptions.
\end{enumerate}

\noindent \textbf{Implementation Details}
For a fair comparison with prior work, we initialize our modality-specific encoders using the pre-trained CLIP-ViT/B-16 \cite{clip} model, the same version used by IRRA \cite{irra}. To increase data diversity, we apply random horizontal flipping, random cropping, and erasing for images, along with random masking, replacement, and removal for text tokens. Input images are resized to 384 × 128, and the maximum text length is set to 77 tokens. We train the model for 60 epochs using the Adam optimizer with a learning rate initialized to $1\times10^{-5}$ and a cosine learning rate scheduler. The batch size is 256 and temperature parameter $\tau$ is set to 0.015. The hyperparameter for MARS is Middle and Late Layer and the discard ratio is 0.25. Normalization Strategy is L1 Normalization. For ATS, p\_start and p\_end value are 0.65 and 0.5, respectively. We set t\_small value equal to the time when the epoch that baseline achieves the best results.

\section{More Quantitative Results}
To further validate our approach, we conduct extensive quantitative experiments and ablation studies. We analyze the impact of our Adaptive Token Scheduler (ATS), demonstrating in \textbf{\cref{tab:ats}} that a step-level application yields the best results. We also assess the generalizability of our method with different CLIP backbones in \textbf{\cref{tab:backbone}}, confirming consistent performance gains over the baseline. Finally, we provide a detailed comparison of our MARS selection strategy against several heuristic-based alternatives in \textbf{\cref{tab:mars}}, which confirms the superiority of our proposed method.

\begin{table}[h]
    \centering
\begin{tabular}{c|c|c|c|c}
\toprule
 Setting   & R@1 & R@5 & R@10 & mAP \\
    \midrule

    Baseline & 65.00 & 85.45 & 90.15 & 52.26
 \\
  Epoch (ATS)    & 65.25 & 83.85 & 89.95 & 51.39 
  \\

 Step (ATS) & \textbf{65.95} & \textbf{85.70} & 90.10 & \textbf{52.71}
  \\
 \bottomrule
\end{tabular}
    \caption{Ablation study on the effect of the Adaptive Token Scheduler (ATS). The baseline does not use the scheduler, while ATS is applied either at the epoch or step level. The results show that step-level scheduling achieves the best performance, yielding improvements in both R@1, R@5 and mAP compared to the baseline.}
    \label{tab:ats}
\end{table}

\begin{table}[h]
    \centering
\begin{tabular}{c|c|c|c|c}
\toprule
  Setting  & R@1 & R@5 & R@10 & mAP \\
\midrule
\multicolumn{5}{l}{using ViT/B-16 CLIP as backbone} \\
\midrule
Baseline & 61.30 & 80.85 & 87.65 & 49.12 \\
\textbf{Ours} & \textbf{67.30} & \textbf{85.60} & \textbf{90.50} & \textbf{53.05} \\
\midrule
\multicolumn{5}{l}{using ViT/B-32 CLIP as backbone} \\
\midrule
Baseline & 59.65 & 79.35 & 86.35 & 47.40 \\
\textbf{Ours} & \textbf{64.50} & \textbf{84.10} & \textbf{90.40} & \textbf{50.28} \\
\bottomrule
\end{tabular}
    \caption{Ablation study on different CLIP backbones. Our method consistently improves performance over the baseline when applied to both ViT/B-16 and the lightweight ViT/B-32 backbone. Notably, even with the smaller ViT/B-32, our approach achieves clear gains in R@1 and mAP, demonstrating its effectiveness across different model capacities.}
    \label{tab:backbone}
\end{table}

\begin{table}[htp]
    \centering
    \begin{tabular}{c|c|c|c|c}
    \toprule
    Setting & R@1 & R@5 & R@10 & mAP  \\
    \midrule
        A & 60.25 & 79.70 & 88.10 & 48.27 \\
        B & 66.25 & 84.45 & 90.20 & 52.29 \\
        C & 60.95 & 81.40 & 87.60 & 48.99 \\
        D & 65.65 & 84.00 & 89.95 & 51.58 \\
        MARS & \textbf{66.95} & \textbf{85.15} & \textbf{90.40} & \textbf{53.05} \\
        \bottomrule
    \end{tabular}
    \caption{Ablation study of different strategies for selecting top-K patches based on attention statistics. We compare our method (MARS) against four baseline strategies: selecting patches with the minimum mean attention (A), maximum mean attention (B), minimum standard deviation of attention (C), and maximum standard deviation of attention (D). The results clearly demonstrate that our MARS method outperforms all baseline approaches across every evaluation metric. MARS achieves the highest performance with R@1 of 66.95\% and mAP of 53.05\%. This underscores the effectiveness of our selection strategy compared to simpler heuristics based only on the mean or standard deviation of attention scores.}
    \label{tab:mars}
\end{table}


\section{More Qualitative Results}
\label{sec:intro}
\begin{figure*}[p]
    \centering
    \includegraphics[width=\textwidth]{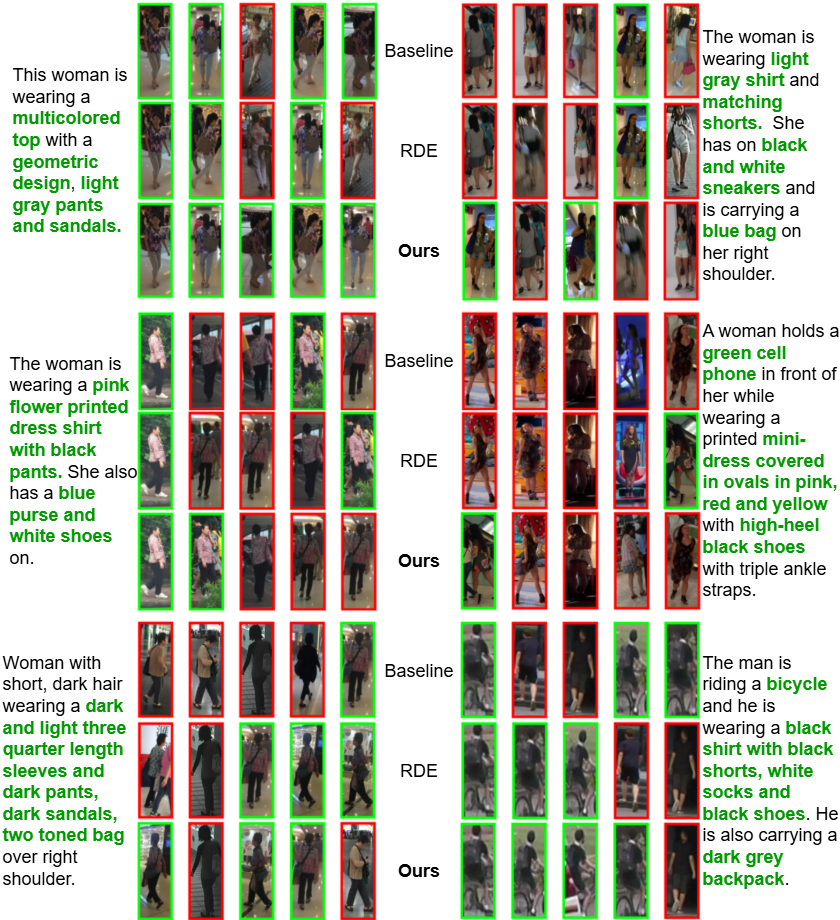}
    \caption{More examples on CUHK-PEDES benchmark}
    \label{fig:cuhk}
\end{figure*}

\begin{figure*}[p]
    \centering
    \includegraphics[scale=0.66]{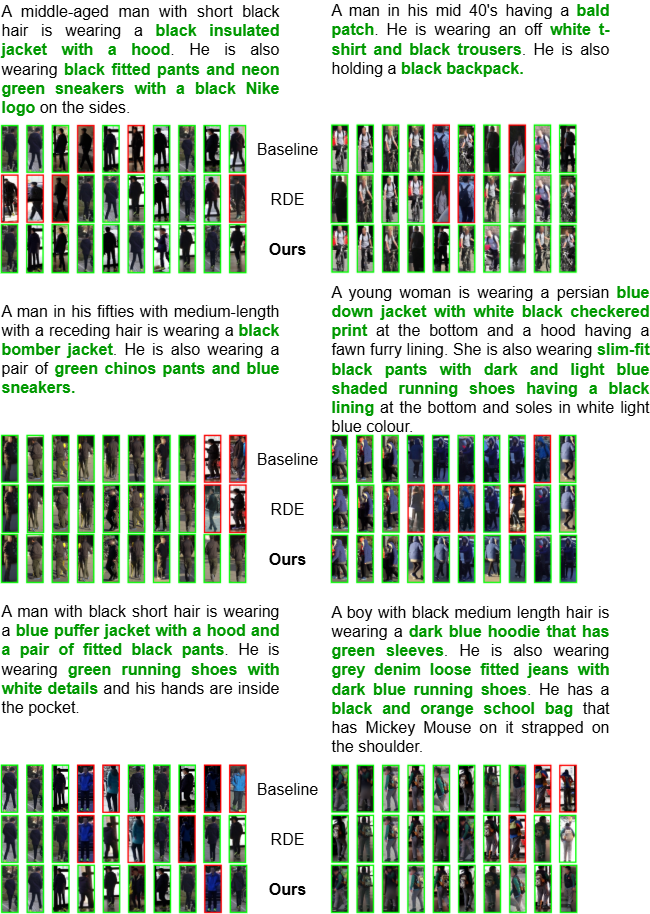}
    \caption{More examples on ICFG-PEDES benchmark}
    \label{fig:icfg}
\end{figure*}

\begin{figure*}[p]
    \centering
    \includegraphics[width=\textwidth]{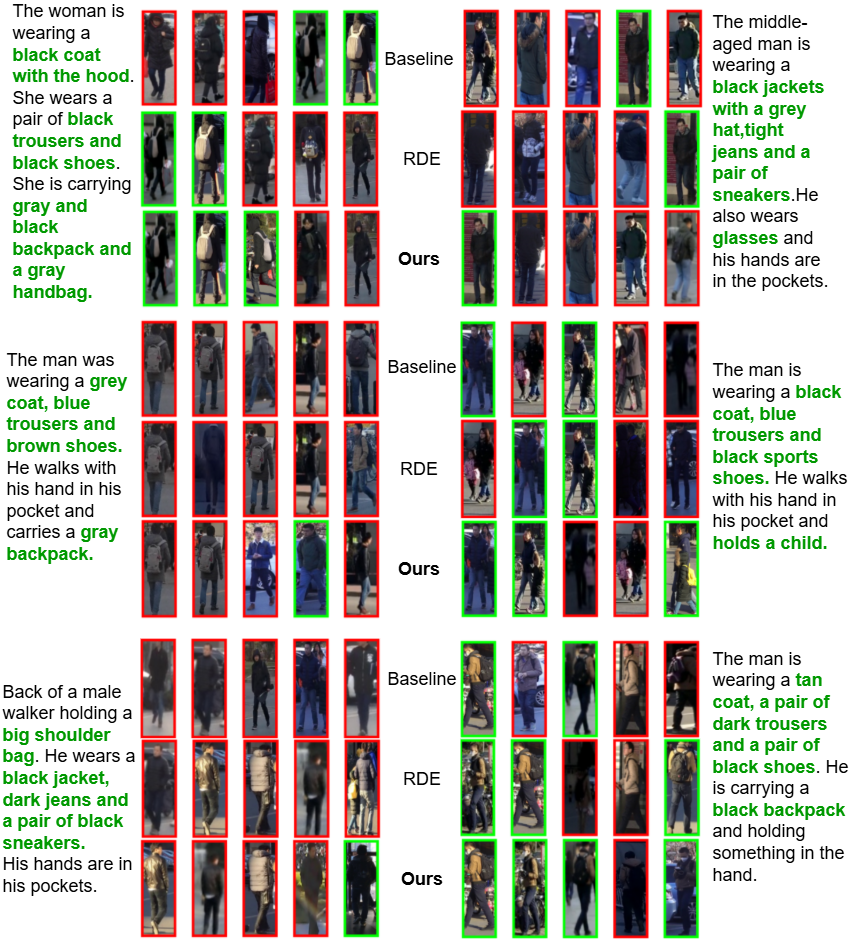}
    \caption{More examples on RSTPReid benchmark}
    \label{fig:rstp}
\end{figure*}

\begin{figure*}[p]
    \centering
    \includegraphics[width=\textwidth]{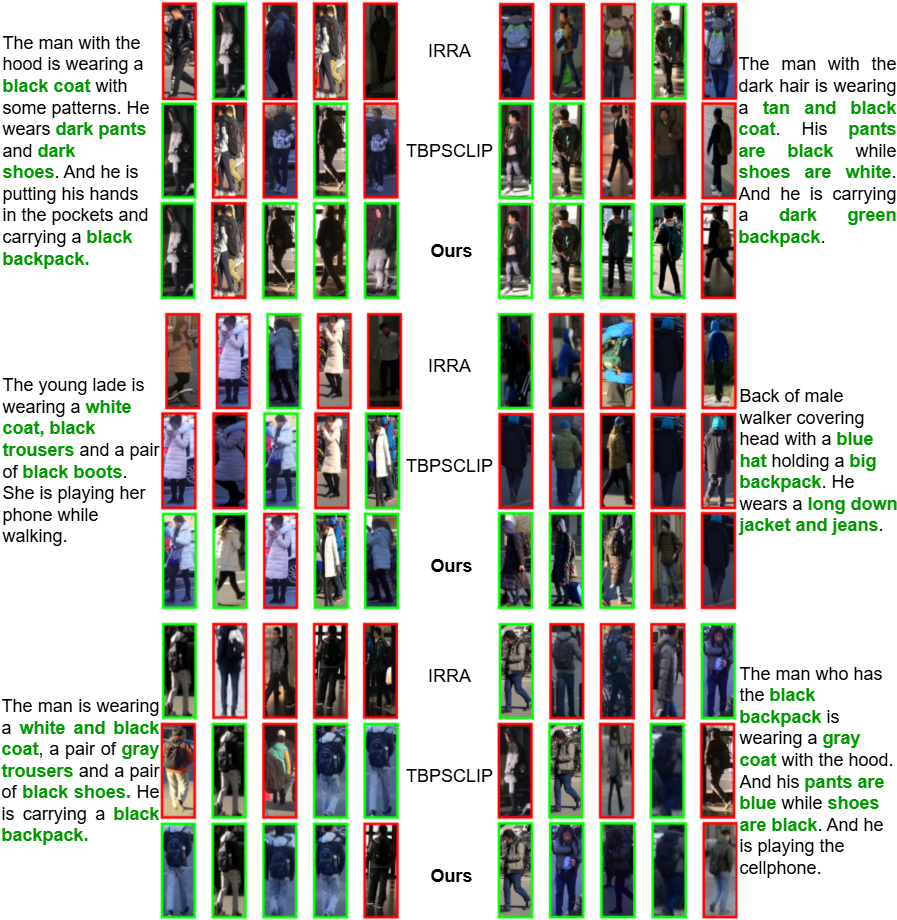}
    \caption{More examples compare with other methods on RSTPReid benchmark}
    \label{fig:extra}
\end{figure*}

\textbf{More Retrieval Results.} \textbf{\cref{fig:cuhk}, \cref{fig:icfg}, \cref{fig:rstp} and \cref{fig:extra}} provide additional qualitative comparisons across the CUHK-PEDES, ICFG-PEDES, and RSTPReid benchmarks. The examples consistently demonstrate that our method retrieves visually and semantically accurate matches, even in challenging cases involving fine-grained attributes, small accessories, and visually similar distractors. Compared to the baseline, RDE \cite{rde}, and other strong methods such as IRRA \cite{irra} and TBPSCLIP \cite{tbpsclip}, our approach shows superior robustness in capturing subtle cues like clothing textures, color combinations, and carried objects (e.g., backpacks, purses, or phones). Notably, our model remains reliable under domain shifts, handling diverse scenarios ranging from crowded street scenes to low-light images. These results further validate the effectiveness of our framework in producing more discriminative and generalizable text-image alignments.

\noindent \textbf{More Attention Map Visualization.}
\textbf{\cref{fig:attn}} presents additional qualitative comparisons of attention maps between our method and RDE on the RSTPReid benchmark. The results show that our model consistently attends to more discriminative and semantically relevant regions described in the text queries, such as specific clothing colors, accessories, and carried items (e.g., backpacks, bags, or bicycles). In contrast, RDE often produces diffuse or misaligned attention, failing to capture fine-grained cues. These visualizations further highlight the effectiveness of our approach in leveraging textual guidance to localize meaningful visual regions, thereby enabling more accurate text-based person retrieval.

\begin{figure*}[p]
    \centering
    \includegraphics[width=\textwidth]{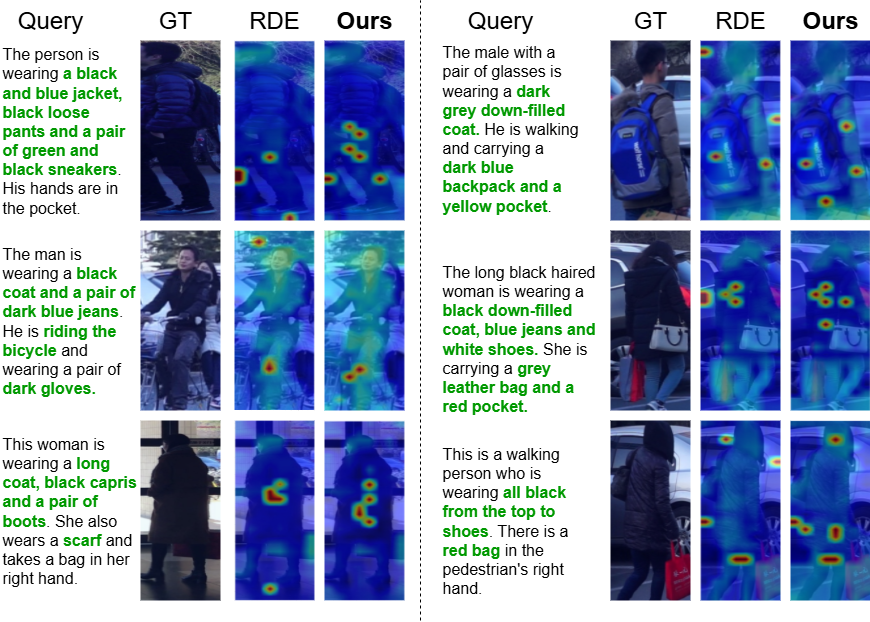}
    \caption{More attention map examples compared with RDE on the RSTPReid benchmark}
    \label{fig:attn}
\end{figure*}

{
    \small
    \bibliographystyle{ieeenat_fullname}
    \bibliography{main}
}







\end{document}